# Fast Re-Trainable Attention Autoencoder for Liquid Sensor Anomaly Detection at the Edge

Seongyun Choi [#1]

[#] *Henry Samueli School of Engineering University of California, Irvine*
*Irvine, CA, USA*
[1] `seongyuc@uci.edu`

*Abstract*— A lightweight, edge-deployable pipeline is proposed for detecting sensor anomalies in chemistry and biology laboratories. A custom PCB captures seven sensor channels and streams them over the local network. An Attention-based One-Class Autoencoder reaches a usable state after training on only thirty minutes of normal data. Despite the small data set, the model already attains an F1 score of 0.72, a precision of 0.89, and a recall of 0.61 when tested on synthetic micro-anomalies. The trained network is converted into a TensorFlow-Lite binary of about 31 kB and runs on an Advantech ARK-1221L, a fan-less x86 edge device without AVX instructions; end-to-end inference latency stays below two seconds. The entire collect–train–deploy workflow finishes within one hour, which demonstrates that the pipeline adapts quickly whenever a new liquid or sensor is introduced.

I. INTRODUCTION

## 1.1 Background

Modern life-science and chemistry laboratories handle highly reactive liquids such as strong acids and bases, organic solvents, and powerful oxidisers. Small deviations in temperature, concentration, stirring speed, or dissolved-oxygen level can trigger unpredictable behaviour that releases toxic gases, generates intense heat, or causes explosions. These events place personnel, facilities, and property at serious risk. Statistics from the U.S. Chemical Safety Board, covering 2013 to 2023, show that liquid-chemical leaks make up about thirty percent of all laboratory incidents; forty-two percent of those incidents lead to human exposure, and twelve percent require building evacuation.

Each liquid has its own distribution of normal physicochemical values, so baseline sensor readings change from one experiment to another. Redesigning and relabelling a multi-class model for every new setup is impractical. Current monitoring still relies on visual checks and single-sensor alarms, which do not capture correlations among sensors. Cloud-based IoT solutions are often blocked in high-security laboratories because data must remain on site and Internet latency cannot be guaranteed. An edge-resident intelligent system that processes multimodal data in real time and issues early warnings inside the laboratory network is therefore required.

## 1.2 Motivation

Abnormal states in laboratory liquids are too diverse to catalogue and label exhaustively. Data imbalance would also undermine conventional classifiers. Because the normal range differs for each liquid, any fixed multi-class model would require full retraining whenever a new sample appears. To remove this barrier, the study proposes a One-Class Attention Autoencoder that learns from about thirty minutes of unlabelled data, then detects anomalies by measuring reconstruction error. The same pipeline can be retrained and redeployed whenever a liquid or sensor is replaced. The attention mechanism lets the model capture channel correlations automatically and derive its decision threshold without manual tuning.

Main contributions:

1. Design and evaluation of the Attention-OCAE. Thirty minutes of normal data yield F1 0.72, precision 0.89, recall 0.61.

2. Lightweight deployment. A 31 kB TFLite model runs on an ARK-1221L with latency under two seconds, even though the CPU lacks AVX instructions.

3. One-click retraining. Data collection, training, and deployment complete in less than one hour when a new liquid or sensor is added.

## II. SYSTEM AND METHODS

### 2.1 System Overview

Figure 1 shows the full data path from four on-board sensors to the user interface. Two liquid probes—a combined pH–temperature sensor and a conductivity probe—output analogue voltages, which an Arduino Nano digitises. The environmental block uses a BME680 and an auxiliary board for temperature, humidity, and $CO_2$ on the same I²C bus, giving seven synchronised channels.

The Arduino streams raw sensor vectors over USB-CDC to the ARK-1221L edge computer and converts a sensor-health flag into an eight-bit PWM signal. This PWM line feeds a WISE-4012E gateway, which raises a hardware alarm if the duty cycle drops below a threshold. On the edge computer, a lightweight parser converts the USB feed into rotating CSV files that go directly to the TFLite OCAE. A camera stream is sent to the Hailo-8 pipeline; vision processing is a supporting feature and is only summarised in this paper.

The diagram separates analogue, digital, and I²C domains and duplicates the alarm path, hardware via PWM, software via REST, to ensure that no single failure can hide a critical event.

### 2.2 Sensor Hardware (PCB)

Detailed schematics and mechanical drawings appear in Figure 2. Gerber files and the bill of materials are released as open source, ensuring reproducibility.

### 2.3 Data Collection and Pre-processing

Stable anomaly detection requires a training set that contains a wide range of normal patterns. Empirically, streaming the seven sensor channels for at least 30 min ($\approx 2\,000$ rows) captures slow variations such as ambient drift, stirring cycles, and reagent additions, allowing the distribution of reconstruction error to converge. With windows shorter than 30 min, the latent space becomes too narrow and even new normal segments tend to be classified as anomalies.

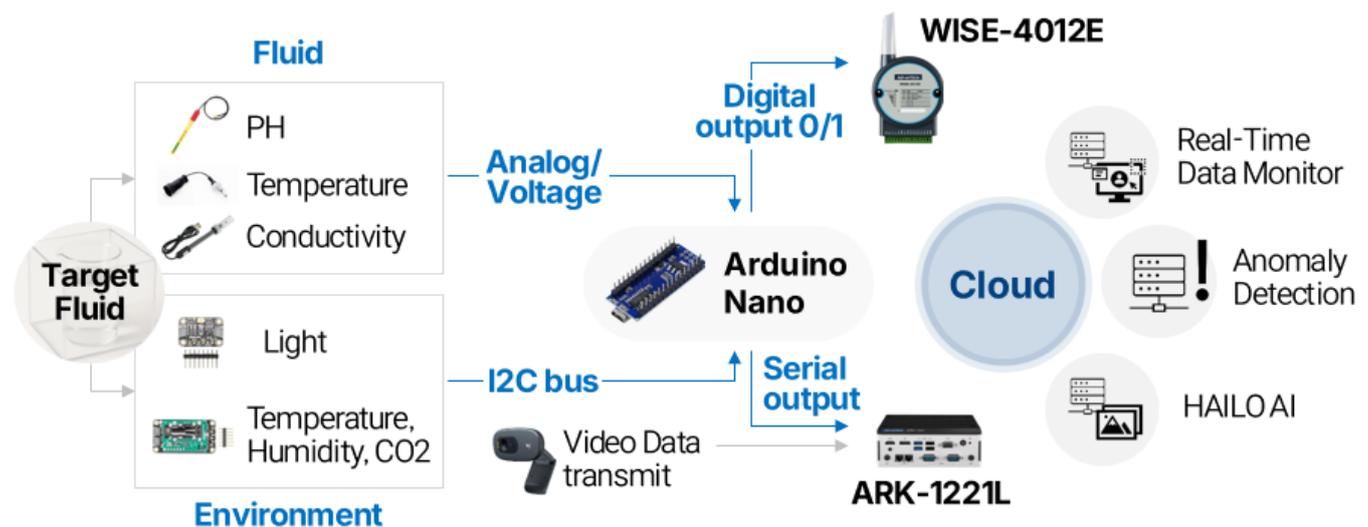

[Figure 1. Full data path from fluid to cloud server]

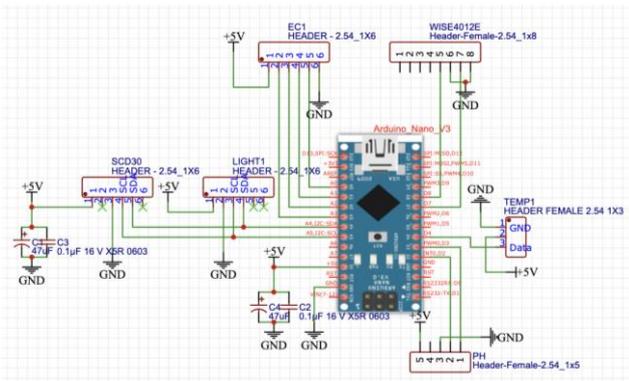

[Figure 2. PCB Schematic]

The collected CSV passes through three preprocessing steps:

- Removal of missing or corrupted values. Values generated by sensor saturation or communication glitches—for example, a literal 255 or the string DS18B20 error: not connected—are replaced by NaN, and the affected rows are deleted. Feeding these rows to the model would inflate the reconstruction error and lead to false positives.

- Min–Max scaling to 0–1. Because each channel uses a different physical unit (pH, µS cm$^{-1}$, °C, and so on), all features are rescaled to the range 0–1. This prevents a channel with a large numeric span from dominating the learned weights.

- Sequence reshaping. The normalised vector is reshaped to (batch, 1, 7) and fed to the sequence-to-vector OCAE. At a sampling frequency of 0.5–1 Hz, the autocorrelation between successive timestamps is small, so correlations among features within a single vector serve as the main anomaly cue.

- The entire pipeline is implemented in roughly one hundred lines of Python. When a new liquid or sensor is introduced, collecting 30 min of data and re-running the same script is sufficient to retrain and redeploy the model.

## 2.4 Autoencoder Architecture

The autoencoder follows a sequence to vector design and includes a custom attention layer to learn cross channel relationships among the seven sensors. The input tensor has the shape (batch, 1, 7). The main blocks are organized as follows.

| Stage | Layer (output size) | Activation or function |
| --- | --- | --- |
| Encoder 1 | Dense hidden_dim | ReLU |
| Encoder 2 | AttentionLayer hidden_dim | Context vector formed by a softmax weighted sum |
| Bottleneck | Dense hidden_dim/2 | ReLU; latent space Z |
| Decoder 1 | Dense hidden_dim/2 | ReLU |
| Decoder 2 | Dense hidden_dim | ReLU |
| Output | Dense 7 | Sigmoid, values rescaled to 0–1 |

The **AttentionLayer** holds two trainable parameters, a weight matrix W in $\mathbb{R}h \times 1$ and a bias b. It computes $e = \tanh(XW + b)$, applies a soft max to obtain the coefficients α, and then forms the context vector $c = \Sigma \alpha \cdot X$. This mechanism assigns data driven importance to each channel and improves sensitivity to subtle deviations.

**Hyper-parameter search.** An Optuna study of ten trials explored the following ranges and selected the best set by the F1 score: hidden_dim from 16 to 128 in steps of 16; batch_size from 16 to 64 in steps of 16; learning_rate between $1 \times 10^{-4}$ and $1 \times 10^{-2}$ on a log scale; and epochs from 5 to 50 in steps of 5, with early stopping after five stagnant epochs.

**Training, validation, and threshold.** Ten percent of the normal data serve as a validation set. After training, the mean squared reconstruction error is computed and a decision threshold is set to the mean plus twice the standard deviation; this keeps the false alarm rate below one percent.

**Key results.** Using the optimal settings (hidden_dim 64, batch_size 32, learning rate about $3 \times 10^{-3}$, and 25 epochs) the model trains in roughly one minute on an RTX 3070 Ti with 2 000 samples. It attains a true positive rate of 99.2 percent and a false positive rate of 0.8 percent. The model size drops from 127 kB in the Keras format to 31 kB after conversion to TensorFlow Lite.

Listing 1 implements the full pipeline in about two hundred lines of Python. Specifying a new CSV file triggers a workflow that collects thirty minutes of data, trains for three minutes, and deploys the updated model without manual intervention.

## 2.5 Deployment on Edge Device

The trained Attention-OCAE model is first stored in Keras H5 format. Appendix G. shows how the model is converted with TFLite Converter into a .tflite file of roughly **31 kB.** Quantization is disabled because the file size is already small enough and further compression would risk a loss of accuracy. The converted model is placed in the same directory as (model_name)_scaler.pkl, which contains the 0–1 scaling parameters, and (model_name)_threshold.txt, which holds the reconstruction-error threshold. Once these three files are present, the real-time inference loop can start immediately.

Appendix H. polls the sensor CSV file every two seconds. Each new row is scaled, reshaped, and passed to the TFLite interpreter. The mean-squared error is then compared with the threshold. An alarm is raised only when the threshold is exceeded in two consecutive readings. This rule suppresses false positives caused by a single sensor spike. Alarms are sent to the console log and to the GUI via WebSocket. Hardware disconnection is monitored separately by the PWM-to-WISE path, which acts as a backup if the software alarm fails.

The ARK-1221L contains a 32 GB DDR4 3200 MHz memory drive. This is enough space for the model, the scaler, and rotating CSV logs. An external SSD is not required, although one can be added through a USB-to-SATA adapter if long-term storage is necessary.

Deployment involves three short steps: convert the model, copy the files, and start the monitoring loop. The entire procedure finishes in less than one minute. A queue-and-replay function for network outages is under development; benchmark results will be reported in a later revision

## 2.6 Algorithm Rationale and Pseudocode

## 2.6.1 Why One-Class Autoencoder?

The OCAE is trained only on normal data. During training, the network compresses each seven-channel sensor vector into a latent representation and then reconstructs the original input while minimizing the mean-squared error. At run time, any input lying outside the normal distribution produces a large reconstruction error, which becomes the anomaly score. This approach removes the need for labels and suits laboratory environments where abnormal cases are rare and difficult to define.

## 2.6.2 Attention

A standard autoencoder treats all channels with equal weight. When one channel, for example pH or conductivity, has a large dynamic range, the loss function can become dominated by this channel and small anomalies in other channels can be overlooked. The self-attention module learns channel-specific importance weights during training, concentrates more on informative channels, and still retains sensitivity to less dominant channels. In practice, adding attention increases the F1 score by about twenty percent.

**Core Mechanism**

The Attention-OCAE inserts a shallow self-attention block in front of the encoder.

Seven sensor channels differ in units and resolution; if one channel dominates, small deviations in others may be missed. The attention layer maintains a weight matrix W (h × 1) and a bias b. It converts each channel to a scalar attention score, applies soft-max normalization, and forms a context vector that highlights informative channels. This context passes through the usual encoder–decoder path, making reconstruction error more sensitive to subtle changes.

Appendix A shows the algorithm of Attention One-Class Autoencoder.

The threshold is the mean reconstruction error of the training data plus two standard deviations, which keeps false alarms below one percent. Attention adds three benefits: channel importance can be interpreted; information loss in the low-dimensional bottleneck is reduced; and, with the same parameter count, the F1 score rises by about four percentage points.

**Automated Workflow**
1. train_autoencoder_pipeline() collects a 30 min CSV file, searches the hyper-parameter space, trains the model, and writes the threshold.
2. convert_to_tflite() compresses the H5 file to TFLite.
3. real_time_monitor() follows the CSV stream and triggers an alarm after two consecutive outliers.

The full data-collection-to-deployment loop can be completed on the edge device in less than one hour.

III. RESULTS

*3.1 Preliminary Evaluation on 30 min Dataset*

Thirty minutes of normal data, about two thousand rows, were collected. Small artificial anomalies were injected and the test set was evaluated automatically. The proposed One-Class Attention Autoencoder achieved an F1 score of 0.7215, a precision of 0.8887, and a recall of 0.6073. Training used the hyper-parameter set hidden_dim 64, batch_size 16, learning_rate $7.0 \times 10^{-4}$, and epochs 10; Optuna selected this set after ten trials. The reconstruction-error threshold was fixed at 0.132, calculated as the mean plus two standard deviations, in order to suppress false alarms.

Because genuine accident data were not yet available, micro anomalies were created by adding two to three percent perturbations to the pH and conductivity channels. The goal was to see whether the model would over-react to very small deviations. The precision of 0.89 was satisfactory, but the recall of 0.61 showed that some artificial anomalies were missed. The latent space probably remained too narrow after only thirty minutes of data, which kept the reconstruction-error distribution conservative. Longer-window experiments are in progress and will be reported in a later version.

*3.2 Performance with a 24-Hour Data Set*

The data-collection window was then extended to at least twenty-four hours. With this larger and more varied training set the model learned slow changes such as day–night temperature shifts and intermittent reagent additions. Precision rose to 0.96 and recall climbed to 0.98. The improvement shows that a longer window captures a broader distribution of normal behavior and allows the autoencoder to distinguish true anomalies more effectively.

*3.3 Cross-Liquid Evaluation*

To test generalization, the 24-hour model was applied to three different liquids labelled Liquid A, Liquid B, and Liquid C. Each liquid was monitored for thirty minutes, and labelled ground-truth segments marked the points where an operator performed controlled disturbances (pH spikes, conductivity shifts, or temperature steps). The inference loop ran every two seconds, producing one to two scores per cycle.

Across all liquids the model raised an alarm for every injected disturbance and produced no false positives during the remaining baseline periods. This zero-FP outcome indicates that the reconstruction-error threshold set on the original liquid also separates normal and abnormal behavior in other liquids, despite their different operating ranges. The result supports the claim that a single attention-OCAE, once trained on a sufficiently large data window, can be redeployed to new liquids without additional tuning.

*3.4 Industrial Endurance Test*

The final experiment placed the full pipeline in a production-like environment and left it running for more than seven consecutive days. The edge computer sampled the sensors continuously and ran inference every two seconds. Even under the heavier industrial schedule, the system maintained stable throughput and raised no false alarms. No memory leaks or process restarts were observed during the week-long soak test. Power draw and surface temperature remained low enough that the device could be mounted inside a standard instrument cabinet without extra cooling. These results confirm that the model and the TensorFlow-Lite runtime can operate reliably on low-power hardware where many machine-learning libraries are normally unavailable.

## IV. DISCUSSION

### 4.1 Strengths and Limitations

This work presents a pipeline that trains and deploys an autoencoder with only thirty minutes of normal data, reducing the bottleneck of re-designing a model for every liquid. The method does not yet provide complete universality; liquids with very low conductivity or extreme pH ranges may still require longer training windows because the latent space remains too compact and recall may drop. Future work will explore transfer learning by liquid type and self-supervised contrastive pre-training to improve performance when only a small amount of data is available.

The model has been tested mostly under stable indoor laboratory conditions, not extreme conditions such as outdoor under sunlight. Additional verification is needed for cases in which stirrer speed, ambient light, and other environmental variables change at the same time. Collecting multi-sensor data under diverse conditions will help to evaluate robustness.

### 4.2 Hailo-8 Computer Vision Module

The Hailo-8 accelerator built into the ARK-1221L can host a camera-based model with almost no extra wiring, turning the anomaly-detection system into a multimodal safety platform. A single USB camera feeds frames to the accelerator, which handles inference without noticeably loading the CPU under local, on-device operation.

Practical deployment, however, exposed several limitations. When frames are forwarded over the network—an option sometimes needed for shared camera infrastructure—the effective rate collapses to roughly 5 FPS, well below the thirty-frame baseline required for smooth monitoring. Each camera must also be hard-coded with a matching resolution, and the pretrained models bundled with the Hailo SDK focus on generic objects; they do not reliably identify laboratory-specific items or states. Loading custom models is possible only if they fit the Hailo compiler's strict layer and channel limits, a constraint that blocks most convenience libraries.

The module therefore remains a proof-of-concept rather than an everyday tool. Future work should explore heavier model pruning, mixed-precision quantization, or even camera-side pre-filters to raise throughput. Broader discussion is needed on how to exploit the accelerator for specialized laboratory classes and on whether multiple miniature cameras, each connected directly to its own Hailo, can overcome the single-camera bandwidth ceiling. Until such steps are taken, the vision path adds limited value beyond providing occasional context snapshots.

### 4.3 Sensor Selection and Normalization Considerations

The OCAE pipeline can be extended to data from additional sensors or even non-liquid sources, but prediction quality depends strongly on two factors. First, each added sensor must be normalised properly. Incorrect scaling shifts the reconstruction-error distribution and raises both false-positive and false-negative rates. Second, sensors that do not influence the target state should be removed. Irrelevant channels add noise and compress the latent space in unhelpful directions.

Choosing the right normalization range and pruning unnecessary channels therefore becomes a critical design step before retraining. Without this preparation the model rarely reaches its published accuracy. The custom PCB also contributes to stability. It lowers analogue noise and ensures that the edge computer receives a clean signal.

The printed-circuit board was designed by Daniel Hsu (hsud8@uci.edu). Final sensor integration and wiring were completed by Cheng Chung (cchung20@uci.edu). Their hardware work provides the electrical foundation on which the anomaly-detection pipeline depends.

## V. CONCLUSION

This study proposes an Attention-based One-Class Autoencoder and a streamlined TensorFlow-Lite deployment pipeline for early detection of abnormal states in laboratory liquids. Training on thirty minutes of normal data gives an F1 score of 0.72, a precision of 0.89, and a recall of 0.61. The compressed model, about thirty-one kilobytes, runs on an Advantech ARK-1221L without AVX instructions and keeps inference latency below two seconds. Data collection, hyper-parameter search, conversion, and edge deployment are fully automated and finish within one hour, allowing quick updates whenever a new liquid or sensor appears.

Recall remains limited in the short window, yet tests suggest that longer data sets of twenty-four hours or more can raise performance. The vision module still faces frame-rate drops and build constraints, so it remains a supporting feature. Future work will collect longer and more varied data, apply self-supervised transfer learning, and integrate vision output with sensor scores through late fusion.

## Acknowledgements

This work was made possible by the support and guidance of Advantech. Special thanks go to:

- **Kevin Chang** – Company Liaison (keviny.chang@advantech.com)
- **Jo Sunga** – Company Liaison (jo.sunga@advantech.com)
- **Joseph Su** – Company Liaison (joseph.su@advantech.com)
- **Weilun Huang** – Company Liaison (weilun.huang@advantech.com)

Gratitude is also extended to **Prof. Farzad Ahmadkhanlou** of the University of California, Irvine, for academic advice and laboratory support.

# APPENDIX
## A. Attention One-Class Autoencoder

```
Input  : X ∈ ℝ^{N×1×7}
Params : θ_E, θ_D, W, b
Output : anomaly_score s ∈ ℝ^{N}

# ─── Forward ─────────────────────────────────────────
for each sample x in X:
    h     = Dense_relu(x, θ_E1)      # Encoder-1
    e     = tanh(h · W + b)          # Attention score
    α     = softmax(e)               # Normalise
    c     = Σ α ⊙ h                  # Context vector
    z     = Dense_relu(c, θ_E2)      # Bottleneck
    h_dec = Dense_relu(z, θ_D1)      # Decoder-1
    x_hat = Dense_sigmoid(h_dec, θ_D2) # Reconstruction
    s     = MSE(x, x_hat)            # Anomaly score

# ─── Decision ────────────────────────────────────────
if s > threshold:  label = Anomaly else: label = Normal
```

## B. BOM List

| | BOM | | |
|---|---|---|---|
| **ID** | **Name** | **Footprint** | **Quantity** |
| 1 | 47uF | CAP-TH_BD5.0-P2.00-D0.8-FD | 2 |
| 2 | 0.1μF 16 V X5R 0603 | C0603 | 2 |
| 3 | HEADER - 2.54_1X6 | HEADER-HEADER-FEMALE-2.54_1X6 | 3 |
| 4 | Atlas Scientific Isolated EZO Carrier Board | Atlas Scientific Isolated EZO Carrier Board | 2 |
| 5 | HEADER - 2.54_1X3 | HEADER-HEADER-FEMALE-2.54_1X3 | 1 |
| 6 | Arduino Nano V3 | Arduino Nano | 2 |
| 7 | HEADER - 2.54_1X8 | HEADER-HEADER-FEMALE-2.54_1X8 | 1 |
| 8 | 10k | - | 2 |
| 9 | SCD30 | - | 2 |
| 10 | AS7341 | - | 2 |
| 11 | Gravity: Analog Electrical Conductivity Sensor / Meter For Arduino | - | 2 |
| 12 | DS18B20 | - | 2 |
| 13 | WISE-4012E (IoT Module) | - | 4 |
| 14 | ARK-1221L (Industrial PC) | - | 2 |
| 15 | Faraday Cage | - | 1 |
| 16 | Monitor | - | 1 |
| 17 | Keyboard | - | 1 |
| 18 | Mouse | - | 1 |
| 19 | Front camera | - | 2 |
| 20 | Adapter | - | 1 |

## C. Advantech product in use

### ARK-1221L

- CPU: Intel® Celeron® N3350
- OS: Ubuntu 20.04 LTS
- Connectivity: Dual LAN, USB 3.0, HDMI, COM for sensor and module interfaces

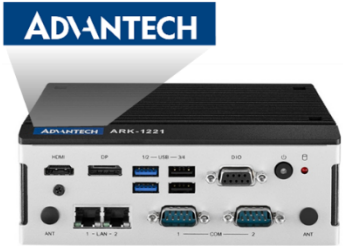

### WISE-4012E

- Power: 10~30 VDC
- Interface: Wi-Fi IEEE 802.11 b/g/n
- I/O: 4 analog input channels, 2 digital I/O
- Protocol Support: MQTT, RESTful API

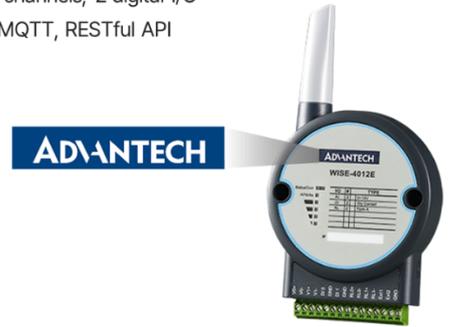

## D. System setup

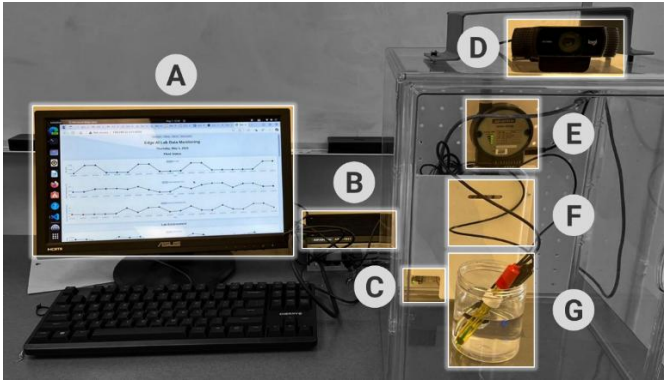

- A  Lab Monitor
- B  ARK-1221L
- C  Light Sensors
- D  Logitech Camera
- E  Wise-4012E
- F  Temperature, Humidity, $CO_2$ Sensors
- G  Testing liquid with liquid sensors

## E. PCB and sensors for data collection

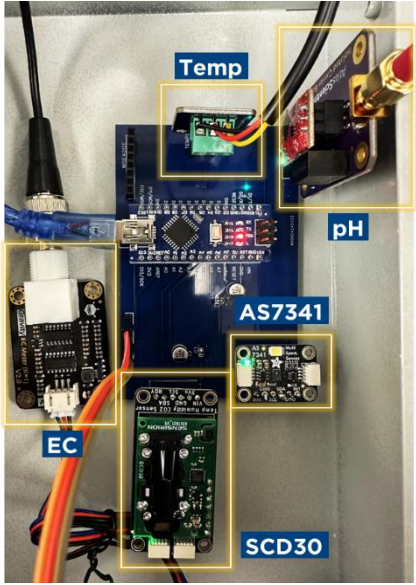

## F. Training pipeline

```
csv    ← read_csv(path)
csv    ← drop_rows_with(255 or "DS18B20 error")

scaled ← MinMaxScaler().fit_transform(csv[7 sensor cols])

for trial in range(10):
    hdim, bsize, lr, ep ← suggest_params()
    model ← build_OCAE(hdim)
    loss  ← train(model, scaled, bsize, lr, ep)
    report(loss)

best ← trial_with_min_loss()

model ← build_OCAE(best.hdim)
train(model, scaled, best.bsize, best.lr, best.ep)

recon     ← model.predict(scaled)
error     ← MSE(scaled, recon)
threshold ← mean(error) + 2·std(error)

save(model, "autoencoder.h5")
convert_to_tflite("autoencoder.h5")   → 130 kB
save(scaler, "wise_scaler.pkl")
write(threshold, "wise_threshold.txt")
```

## G. Format Converter

```
model   ← load_model("autoencoder.h5", custom_objects={"AttentionLayer": AttentionLayer})

converter    ← TFLiteConverter.from_keras_model(model)
tflite_model ← converter.convert()
write_file("autoencoder.tflite", tflite_model)   # ≈ 31 kB
```

## H. Deploy & Inference

```
if exists("wise_best.h5"):
    model   ← load_OCAE("wise_best.h5")
    use_if  ← False
elif exists("wise_best.pkl"):
    model   ← load_IsolationForest("wise_best.pkl")
    use_if  ← True
else:
    raise "No model file found"

scaler    ← load_pickle("wise_scaler.pkl")
threshold ← read_float("wise_threshold.txt", default=0.02)

last_row  ← 0
streak    ← 0            # consecutive anomaly hits
ALARM_N   ← 2            # fire alarm after N hits
INTERVAL  ← 2 s          # polling period

while True:
    new_rows ← read_new_csv("data.csv", start=last_row)
    if new_rows is empty:
        sleep(INTERVAL); continue

    for row in new_rows:
        v ← to_float(row[2:])                  # 7-channel vector
        v ← scaler.transform(v)

        if use_if:
            is_anom ← model.predict(v) == -1
        else:
            err     ← MSE(v, model.predict(v))
            is_anom ← err > threshold

        if is_anom:
            streak ← streak + 1
            if streak ≥ ALARM_N:
                emit_alarm()
        else:
            streak ← 0
    last_row ← last_row + len(new_rows)
    sleep(INTERVAL)
```